\documentclass[conference, onecolumn]{IEEEtran}
\IEEEoverridecommandlockouts
\usepackage{cite}
\usepackage{amsmath,amssymb,amsfonts}
\usepackage{algorithmic}
\usepackage{graphicx}
\usepackage{textcomp}
\usepackage{multirow}
\usepackage{xcolor}
\def\BibTeX{{\rm B\kern-.05em{\sc i\kern-.025em b}\kern-.08em
    T\kern-.1667em\lower.7ex\hbox{E}\kern-.125emX}}
\begin{document}

\title{Unsupervised feature selection for tumor profiles using autoencoders and kernel methods\\
%\thanks{FOCEM, UTN}
}

\author{\IEEEauthorblockN{1\textsuperscript{st} Martin Palazzo}
\IEEEauthorblockA{\textit{Universite de Technologie de Troyes \&} \\
\textit{Instituto de Investigacion en Biomedicina}\\
\textit{de Buenos Aires (IBioBA)—CONICET}\\
\textit{Partner Institute of the Max Planck Society}\\
Troyes, France \\
martin.palazzo@utt.fr}
\and
\IEEEauthorblockN{2\textsuperscript{rd} Pierre Beauseroy}
\IEEEauthorblockA{\textit{Universite de Technologie de Troyes} \\
Troyes, France \\
pierre.beauseroy@utt.fr}
\and
\IEEEauthorblockN{3\textsuperscript{nd} Patricio Yankilevich}
\IEEEauthorblockA{\textit{Instituto de Investigacion en Biomedicina}\\
\textit{de Buenos Aires (IBioBA)—CONICET}\\
\textit{Partner Institute of the Max Planck Society}\\
Buenos Aires, Argentina \\
pyankilevich@ibioba-mpsp-conicet-gov.ar}
}

\maketitle
%\keywords{Kernel Methods, Autoencoders, Cancer genomics, Unsupervised Feature Selection}

\begin{abstract}
Molecular data from tumor profiles is high dimensional. Tumor profiles can be characterized by tens of thousands of gene expression features. Due to the size of the gene expression feature set machine learning methods are exposed to noisy variables and  complexity. Tumor types present heterogeneity and can be subdivided in tumor subtypes. In many cases tumor data does not include tumor subtype labeling thus unsupervised learning methods are necessary for tumor subtype discovery. This work aims to learn meaningful and low dimensional representations of tumor samples and find tumor subtype clusters while keeping biological signatures without using tumor labels. The proposed method named Latent Kernel Feature Selection (LKFS) is an unsupervised approach for gene selection in tumor gene expression profiles. By using Autoencoders a low dimensional and denoised latent space is learned as a target representation to guide a Multiple Kernel Learning model that selects a subset of genes. By using the selected genes a clustering method is used to group samples. In order to evaluate the performance of the proposed unsupervised feature selection method the obtained features and clusters are analyzed by clinical significance. The proposed method has been applied on three tumor datasets which are Brain, Renal and Lung, each one composed by two tumor subtypes. When compared with benchmark unsupervised feature selection methods the results obtained by the proposed method reveal lower redundancy in the selected features and a better clustering performance. 
\end{abstract}

\begin{IEEEkeywords}
Unsupervised Feature Selection, Gene Expression, Kernel Methods, Autoencoders
\end{IEEEkeywords}

\section{Introduction}
\label{sec:introduction}

Cancer informatics has been characterized by the increasing availability of large data repositories like The Cancer Genome Atlas \cite{tomczak2015cancer} and the International Cancer Genome Consortium \cite{zhang2011international}. One of the most used genome technologies to characterize tumor sub-types is Transcriptomics, which measures gene expression \cite{lu2003cancer}. Each tumor profile is described by more than 20.000 gene expression features which defines a high dimensional input space and further complexity in data analysis. Tumor types presents inner heterogeneity that can be sub-divided in tumor sub-types \cite{chen2017glioma} \cite{xiao2017eight} \cite{chen2016multilevel}. 
Given the high dimensional space from the input data it is necessary to reduce the dimensionality while preserving the biological interpretation of the system.  Moreover, tumor labels like tumor subtype or tumor stage are not always available therefore unsupervised approaches that do not need labeled data are necessary in these cases \cite{ang2015supervised}. These reasons motivate us to propose an unsupervised feature selection method for tumor clustering. \\
Feature selection methods are necessary in cancer genomics since they provide a low dimensional representation of the input data. This representation is characterized by a selected subset of the input genes providing interpretability of the results and discarding the rest by following an objective function related to improve a learning task \cite{li2018feature}. In addition, the selected genes can be used to guide biomarker discovery strategies \cite{he2010stable}. The resulting subspace obtained by a feature selection method is described explicitly by a subset of biological features assuming that the initial feature set contains noisy features that can be discarded. A reduced feature subset has benefits in reducing model complexity and in measuring only a reduced set of biological features  \cite{ang2015supervised}.\\
In an unsupervised problem it is expected that the selected genes may improve the clustering of tumor profiles. To guide the feature selection process this work proposes a low dimensional and denoised representation of the input data known as latent space which is learned by an unsupervised neural network known as Autoencoder. Then by a Multiple Kernel Learning process a subset of gene features is selected with the objective to approach as much as possible to the learned representation from the autoencoder. Finally by doing clustering using just the selected features it is expected to observe significant clinical attributes associated to each cluster. Although the latent features learned by autoencoders are useful for further clustering of tumors and subtype discovery, these features are a nonlinear combination of the original ones. For this reason these latent features are not useful for biological interpretation since they are not explicit. 
\\ 
The main contribution of this work is an unsupervised method able to select genes with clinical relevance from high dimensional gene expression data without the need of having tumor labels. The proposed unsupervised feature selection method is applied in tumor data from Lung cancer, Renal cancer and Brain cancer. The performance of the proposed method is compared with two unsupervised feature selection methods. 

\section{Related Work}
Reducing the dimensionality of gene expression data can be achieved by feature selection \cite{ang2015supervised} and feature extraction methods \cite{hira2015review}. \\
Feature extraction is the construction of a reduced subset of new features obtained from a linear or nonlinear combination of the initial set of features. Neural Networks have gained popularity for feature extraction and dimensionality reduction lead by the Autoencoder model which is based on nonlinear transformations \cite{hinton2006reducing}. The reduced dimensional space and the extracted features are known as \textit{latent space and latent features} respectively. Latent features retain salient characteristics from the input data in a low dimensional space \cite{goodfellow2016deep}. Autoencoders have been used in biomedical problems to integrate multi-omic data like gene expression, methylation and microRNA to predict Liver cancer prognosis \cite{chaudhary2018deep}. In a similar way autoencoders have been used to learn meaningful representations from gene expression data of Breast cancer patients and then identify tumor subtypes \cite{guo2019identification}. In addition, Autoencoders have been applied to learn a latent space from somatic mutation data of the pan-cancer landscape showing improvements in the clustering performance \cite{way2017extracting} \cite{palazzo2019pan}. Moreover, Variational Autoencoders (VAE) have been trained on DNA Methylation data from Lung Cancer patients \cite{wang2018exploring} and on gene expression data of pan-cancer tumor samples to learn meaningful representations for supervised and unsupervised tasks \cite{way2017extracting} \cite{gronbech2018scvae}. It is clear the important role and capacity of Autoencoders for feature extraction and dimensionality reduction on molecular data from tumors. \\
Multiple Kernel Learning (MKL) have been used for gene selection in supervised problems with the objective to improve the classification between tumor types \cite{rakotomamonjy2008simplemkl}\cite{du2017feature}. Feature selection has been applied also on multi-omic data like Gene Expression, Methylation and miRNA  using the Minimum redundancy - maximum relevance (mRMR) \cite{ding2005minimum} method to predict survival rate on Glioblastoma Multiforme patients \cite{zhang2016improve}. Another study proposes a stable feature selection method for high-dimensional RNA-seq data while applying an ensemble $L_1$-norm support vector machines to reduce irrelevant features \cite{moon2016stable} and classify tumor stages of renal clear cell carcinoma. In addition, feature selection by Elastic Net \cite{zou2005regularization} has been proposed to select genes linked to the Triple Negative Breast Cancer subtype \cite{lopes2018ensemble}. The papers described above show the potential and necessity of supervised feature selection methods for gene selection on cancer molecular data. \\
Nevertheless, labeled data is not always available and the selection of genes is needed for unsupervised tasks such as clustering since tumor types may present heterogeneity and each cluster can present different clinical properties. This problem is faced by Unsupervised Feature Selection methods for Clustering \cite{alelyani2018feature}. Multi-Cluster Feature Selection (MCFS) \cite{cai2010unsupervised} is an unsupervised model proposed to select the features that preserves the cluster structure of the original data and has been applied on micro-RNA data \cite{bandyopadhyay2015mbstar}. Also the Sparse k-Means (SKM) method \cite{witten2010framework} has been proposed to weight each feature based on the partition of data and by this way a subset of features is selected by penalizing weights with the L1 norm. Moreover, an unsupervised spectral method (SPEC) \cite{zhao2007spectral} has been proposed to determine relevant genes on acute lymphoblastic leukemia \cite{zhao2010integrative}.\\
Feature selection and feature extraction methods both have shown potential to reduce the dimensionality of tumor data. In this work we propose a method that combines both strategies by learning a low dimensional latent space by extracting features from the input data and then selecting the gene expression features that approach to the resulting learned latent representation. 

\section{Materials and Methods}
This work proposes an unsupervised feature selection method based on Kernel Methods guided by the latent structure obtained from an Autoencoder. The method is divided in tree main steps. First build a target kernel matrix composed by the training samples lying on the latent space of an Autoencoder. Then select features by a multiple kernel learning strategy guided by the kernel built in the previous step as the target representation. Finally with the selected features perform clustering and measure the cluster quality by comparing the enrichment of tumor subtypes on each cluster.

\subsection{Datasets}
The proposed method is designed to be used on high dimensional gene expression data from tumor profiles. To evaluate the method three tumor datasets are used: Lung, Renal and Brain cancer. Each type is a separated dataset and is composed by two tumor subtypes samples from the International Cancer Genome Consortium (ICGC) data portal. The Lung Cancer dataset is composed by the projects LUSC-US (Squamus cell subtype) and LUAD-US (Adenocarcinoma subtype) with $478$ and $428$ tumor samples respectively. The Brain cancer dataset is composed by the projects GBM-US (Glioblastoma) and LGG-US (Lower Grade Glioma) with $159$ and $439$ tumor samples respectively. The Renal cancer data is composed by the projects KIRP-US (Papillary) and KIRC-US (Clear cell) with $222$ and $518$ tumor samples respectively. The data matrix for each dataset is $X_{n, d_0}$ with $n$ tumor samples and characterized initially by $d_0 = 17640$ protein coding genes. The objective is to select a subset of genes $p<<d_0$ to reduce the dimensionality.

\begin{table}[h]
\begin{center}
 \begin{tabular}{|c|c|c|c|} 
 \hline
 Type & Subtype & Project & Patients (n) \\
 \hline\hline
 \multirow{2}{2.5em}{Lung Dataset} & Squamus cell & LUSC-US & 478 \\ 
 \cline{2-4}
  & Adenocarcinoma & LUAD-US & 428  \\ 
 \hline\hline
 \multirow{2}{2.5em}{Renal Dataset} & Papillary & KIRP-US & 222  \\
 \cline{2-4}
  & Clear cell & KIRC-US & 518  \\ 
 \hline\hline
 \multirow{2}{2.5em}{Brain Dataset} & Lower Grade Glioma & LGG-US & 439  \\ 
 \cline{2-4}
  & Glioblastoma & GBM-US & 159  \\ 
 \hline
\end{tabular}
\caption{Number of patients by tumor type and subtype.}
\end{center}
\end{table}

\subsubsection{Pre-processing}
To estimate statistically the performance of the proposed method ten independent times $80\%$ of the samples have been randomly selected from the input dataset. The subset of randomly selected samples are used to train the autoencoder and to select the gene features. Each feature has been min-max scaled between 0 and 1 \cite{van2006centering}. Then an univariate filter is applied to reduce the initial number of features from $d_0 = 17640$ to $d = 8820$ by ranking the features by variance and only keeping the $50\%$ best ranked. Univariate filter is used to discard low variance features and perform an initial reduction of the high dimensional space. Nevertheless, the subset of $d$ variables preserved remains large and the sample to feature ratio is $(n/d) = 0.088 $ thus the proposed feature selection method is applied at this point.

\subsection{Autoencoders}
Autoencoders (AEs) are feed forward artificial neural networks (ANNs). AEs have the objective to learn two functions, an encoder and a decoder. The encoder is a non-linear function that maps the input domain $\mathcal{X}$ of $n$ samples and $d$ dimensions to a latent space $\mathcal{Z}$ of lower dimension $l$. On the other side, the decoder is a function designed to reconstruct the samples from the latent space $z$ to the input space. The encoder is forced to learn a function that captures the salient features from $\mathcal{X}$ and maps to $\mathcal{Z}$ \cite{goodfellow2016deep}. The encoder function is defined as $\boldsymbol{z} = f\left ( \boldsymbol{x} \right )$ and the decoder as $\tilde{\boldsymbol{x}} = d \left ( \boldsymbol{z} \right )$. The samples at the latent space are expressed by $\boldsymbol{z}$ while $\tilde{\boldsymbol{x}}$ represents the reconstructed samples by the decoder function lying on $\mathcal{X}$. During training the autoencoder has to minimize following loss function 
\begin{equation}
L\left ( \boldsymbol{x},\tilde{\boldsymbol{x}} \right ) = L\left ( \boldsymbol{x},d\left ( f\left ( \boldsymbol{x} \right ) \right ) \right )
\end{equation}

where $L$ penalizes $d\left ( f\left ( \boldsymbol{x} \right ) \right )$ when it is different from $\boldsymbol{x}$. The loss function is computed by the Mean Squared Error (MSE) expressed as
$$
L_{MSE} (X, \tilde{X})) =  \sum_{i=1}^{n} || x_i -  \tilde{x}_i ||^{2}
$$
Then the encoder $F$ and decoder $D$ functions are expressed as \cite{kampffmeyer2017deep}
$$
\begin{matrix}
\boldsymbol{z} = F\left ( \boldsymbol{x}, \textbf{W}_F \right ) = & \sigma \left ( \textbf{W}_F \boldsymbol{x} + \textbf{b}_F \right )  \\ 
\tilde{\boldsymbol{x}} = D\left ( \boldsymbol{z}, \textbf{W}_D \right ) = & \sigma \left ( \textbf{W}_D \textbf{z} + \textbf{b}_D \right )
\end{matrix}
$$
The encoding function is $F\left ( \cdot , \textbf{W}_F \right )$ and the decoding $D\left ( \cdot, \textbf{W}_D \right ) $. The expression $\sigma \left ( \cdot \right )$ is the activation function of the network. The vectors \textbf{W} and \textbf{b} are the network parameters to learn with the objective to minimize the loss function and represent the weights and biases of the encoder and decoder functions respectively. The optimizer used to learn the parameters of the network is the Adaptive Moment Estimation (Adam) \cite{kingma2014adam} which computes depending on the gradient mean an adaptive learning rate to speed up the learning process.\\
In order to force the encoder to learn a useful representation on the latent space and to avoid the AE to just copy $\tilde{\boldsymbol{x}}$ from $\boldsymbol{x}$ different regularization strategies are implemented. First, a regularization term using the $L_2$ norm is imposed on the weights $W_F$ and added to the loss function ${L}$. The regularization hyper-parameter is $\beta$ as follows
\begin{equation}
 L_{r} = L\left ( \boldsymbol{x},d\left ( f\left ( \boldsymbol{x} \right ) \right ) \right ) + \beta\sum_{i}\left || w_i \right || _2
\end{equation}

The regularization term avoid both $f$ and $d$ to have large weights and leads to learn a simpler model, in consequence this reduces the overfitting of the trained model. A second regularization strategy that helps to improve even more the generalization capacity of the model is Batch Normalization (BN) \cite{ioffe2015batch} which consists to perform normalization at each mini-batch iteration during training.

%\begin{figure}[h]
%  \centering 
%  \includegraphics[width=2in]{autoencoder.png}
%  \caption{A diagram of an autoencoder.}
%  \label{fig:example} 
%\end{figure} 

In this work autoencoders are used to learn meaningful representations from gene expression data from cancer patients. We propose also an architecture for both the Encoder and the Decoder functions. Starting from the Encoder the input layer of dimension $d$ is fully connected to a hidden layer $\textup{HL}_1$ of 200 neurons. Then $\textup{HL}_1$ is fully connected to a second hidden layer $\textup{HL}_2$ with 100 neurons. Finally $\textup{HL}_2$ is connected to the latent hidden layer $\textup{HL}_z$. The latent representation $Z$ has a lower dimension $l = 50$ in comparison to the input dimension $d$ and describes with less noise the original data. Symmetrically and starting from the latent layer $\textup{HL}_z$ the Decoder Function has one hidden layer $\textup{HL}_3$ of 100 neurons followed by  one $HL_4$ of 200 neurons and finally the output layer of dimension $d$. Using the sample set $\{z_i\}$ in $Z$ a Kernel $K_{z}$ is built and used as target kernel by the following Multiple Kernel Learning step. In the next section Kernel Methods are defined and is explained how features are selected to approach the target representation obtained from the autoencoder.

\subsection{Multiple Kernel Learning}

A kernel $k$ is a symmetric function $k:\mathcal{X}\times \mathcal{X} \rightarrow \mathbb{R} $ that meets the condition 

\begin{equation}
k\left ( x_i, x_j \right ) = \left \langle \phi (x_i) , \phi (x_j)  \right \rangle_H
\end{equation}

where $\mathcal{X} \subseteq \mathbb{R}^{d}$ is a d-dimensional space $\mathcal{X}$, and $\phi$ is a mapping function from $\mathcal{X}$ to a high dimensional feature Hilbert space $\mathcal{H}$ with a dot product such that

\begin{equation}
\phi : \mathcal{X} \mapsto \phi \left ( \mathcal{X} \right ) \in \mathcal{H}
\end{equation}

and $\mathcal{H}$ is a Reproducing Kernel Hilbert Space (RKHS). Considering a set of $n$ samples such that $\left ( x_1, x_2, ..., x_n\right ) $ from $\mathcal{X}$ the Kernel or Gram Matrix $\mathbf{K}$ is a $n \times n$ matrix with entries defined as $K_{ij} = \left \langle x_i,x_j \right \rangle$. Each position of the Kernel Matrix is expressed as

\begin{equation}
K_{ij} = \left \langle \phi \left ( x_i \right ), \phi \left ( x_j \right ) \right \rangle = k\left ( x_i, x_j \right )
\end{equation}

and the kernel function is symmetric and positive semi-definite. By using Kernel functions is not necessary to compute explicitly the mapping $\phi$ and the dot products between a pair of samples in the RKHS are computed directly with $k\left ( x_i, x_j \right )$. A Kernel can be thought as a similarity function between vectors samples where at the presence of a pair of orthogonal vectors outputs $0$ and at the presence of similar or equal vectors outputs a positive value thus $K_{ij} \in [0,1]$.  This work uses the Gaussian Kernel defined as

\begin{equation}
k(x_i, x_j) = \textup{exp}\left ( \frac{\left \| x_i-x_j \right \|^{2}}{2\sigma ^{2}} \right ); \sigma > 0
\end{equation}

where $\sigma$ defines the bandwith of the kernel function. The value of $\sigma$ parameter is obtained by computing the median of the pairwise distances of the tumor data \cite{gretton2005measuring}.\\
The Alignment $A(K_1, K_2)$ between two valid kernels $K_1$ and $K_2$ built from a set of samples $n$ is defined as

\begin{equation}
\mathit{A}\left ( K_1, K_2 \right ) = \frac{\left \langle K_1, K_2 \right  \rangle_F }{\sqrt{\left \langle K_1, K_1 \right  \rangle_F \left \langle K_2, K_2 \right  \rangle_F}}
\end{equation}

and measures how close both kernels are in terms of finding similarities between pair of samples from $n$ \cite{cristianini2002kernel}\cite{kandola2002extensions}. The operation $\left \langle K_1, K_1 \right  \rangle_F$ means the Frobenious inner product between the two kernel matrices.\\
Kernels can be combined to form more complex functions capable of handling the biological data. Multiple Kernel Learning (MKL) builds a kernel $k_{\mu }$ from a linear combination of single kernels $k_i$ \cite{gonen2011multiple} and is expressed as

\begin{equation}
\boldsymbol{\mathbf{K}_{\mu }}\left ( \boldsymbol{x},\boldsymbol{x{}'} \right ) = \sum_{i=1}^{n} \mu _i \mathbf{K}_i\left ( \boldsymbol{x},\boldsymbol{x{}'} \right ), \mu _i \geq 0
\end{equation}

The vector $\mu$ represents the weight $\mu_i$ of each kernel $k_i$ and reflects the relative importance of each kernel in the final solution $\mathbf{K}_{\mu }$. There are many objective functions to optimize while learning the MKL task, nevertheless in this work the MKL model is built with the objective to maximize the alignment of the resulting kernel $\mathbf{K}_{\mu }$ and a target kernel $\mathbf{K}_t$. Particularly, the target kernel proposed in this method is the one $\mathbf{K}_{z}$ built from the samples $z_i$ lying at the latent space $\mathcal{Z}$ of the Autoencoder and the resulting alignment to optimize is $A(\mathbf{K}_{\mu }, \mathbf{K}_{z})$.\\
To compute the vector of weights $\mu$ and the resulting  $\mathbf{K}_{\mu}$ a greedy strategy is used \cite{pothin2006greedy}. This approach combines just two kernels $k_1$ and $k_2$ at each iteration while maximizing the aligment $A(\mathbf{K}_{\mu }, \mathbf{K}_{z})$ of the resulting kernel. At each iteration the weights $\mu_1$ and $\mu_2$ of the resulting kernel $\mathbf{K}_{\mu}$ are obtained by computing the derivative of the alignment and find where it becomes $=0$ as optimal condition for each partial derivative. If $\boldsymbol{\mu} = \left [ \mu_1,\mu_2  \right ]$ at each iteration then $\mu_1$ and $\mu_2$ are obtained from 

$$
\begin{matrix}
\frac{\partial \boldsymbol{A}\left ( \mu_1,\mu_2 \right )}{\partial \mu_1} =0 & & 
\frac{\partial \boldsymbol{A}\left ( \mu_1,\mu_2 \right )}{\partial \mu_2} =0 
\end{matrix}
$$

The condition $\mu_1,\mu_2 \geq  0$ is needed since only positive linear combination of kernels are valid kernels. Starting from a set of kernels $D = [K_1,K_2,...,K_d]$ the greedy MKL algorithm chooses at the first iteration the kernel $K_i$ with highest $A(K_i, K_{z})$. Then it will start adding at each iteration a new kernel to the solution that improves as much as possible the current alignment until $A(K_i, K_{z})$ stop increasing. The final vector $\mu$ is sparse with only $\mu_i > 0$ for every $K_i$ selected during the MKL process.

\subsection{Proposed Latent Kernel Feature Selection Method}
Given a set of $n$ tumor samples characterized by $d$ gene expression features the data is contained in a $X_{nd}$ expression matrix. It is desired to select a subset of $p$ features from $d$. To achieve this goal first an autoencoder model is trained and a latent space $\mathcal{Z}$ of dimension $l$ is obtained where $l << d$. Then using the set of samples $n$ projected in the latent space a gaussian kernel $K_{z}$ is built and used as the target kernel. Secondly from $X_{nd}$ a set of $d$ feature-wise kernels are built producing one kernel per feature. Finally by using the MKL model described previously a reduced subset of $p$  kernels is iterativelly selected and combined to build a $K_{\mu}$ kernel that increases the alignment $A(K_{\mu}, K_{z})$. 

\begin{figure}[h]
  \centering
   \includegraphics[width=4.1in]{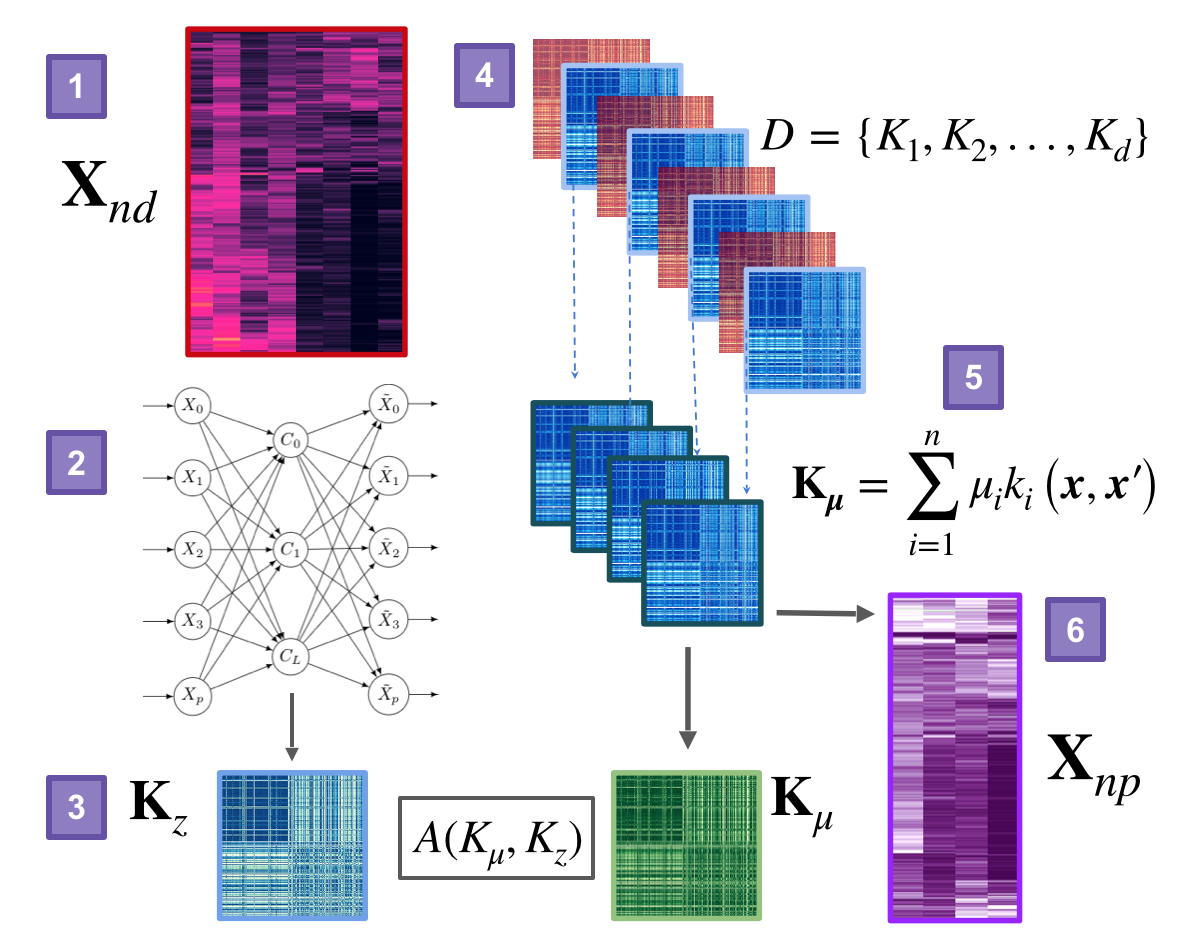} 
  \caption{Pipeline of the proposed method. First starting from the raw data (1) an autoencoder is trained (2) and a latent space learned. Then a $K_{z}$ kernel built (3) using the sample set projected on the latent space. Finally feature-wise kernels are built (4) and combined by MKL (5) to obtain a $K_{\mu}$ kernel by improving the aligment $A(K_{\mu},K_{z})$. The result is a $X_{np}$ matrix characterized by a subset of $p$ features associated to the feature-wise kernels selected by MKL.}
  \label{fig:example1}
 \end{figure}

Only the feature-wise kernels that increases the alignment $A(K_{\mu}, K_{z})$ are included in the final kernel $K_{\mu}$. This approach leads to a sparse solution where the non-zero values of the $\mu$ vector indicates the feature importance on the result. Features are selected by an unsupervised strategy that best align the representation learned from the autoencoder. We name this method Latent Kernel Feature Selection (LKFS). Figure 1 shows a diagram of the proposed method.

\subsection{Evaluation of selected features}

To evaluate the quality of the selected features the Redundancy Rate (RED) \cite{zhao2010efficient} \cite{yamada2014high} metric is used in this work. This metric measures between the selected features the mean value of absolute correlation and is computed as
\begin{equation}
\text{RED} = \frac{1}{p(p-1)}\sum_{f_i, f_j \in \textbf{P}} |\rho_{ij}|
\end{equation}

where $\rho_{ij}$ is the correlation index between the \textit{ith} and the \textit{jth} selected features. The RED score takes values between $0$ and $1$. A RED value close to $0$ means low linear correlation between the selected features thus a low redundancy which is a desired result. On the other side when the values of RED are close to $1$ means that the selected features are highly redundant which is a non desired output. \\
To evaluate visually the quality of the selected features it is possible to use a non-linear dimensionality reduction method that projects the tumor samples characterized by the selected $p$ features in a two-dimensional representation used only for visualization purposes. Then it is possible to evaluate how the tumor samples are distributed in the new feature subset. The t-distributed stochastic neighbor embedding (t-SNE) \cite{maaten2008visualizing} is used in this work for this purposes. \\
Another way to evaluate the quality of the selected features is to perform a unsupervised learning task like clustering after feature selection and measure the quality of the obtained clusters. K-means clustering method \cite{jain2010data} is applied on the tumor samples characterized by the selected $p$ gene features and each tumor sample $x_i$ is assigned to a cluster which implies to assign a cluster label $c_i$. The quality of clusters is evaluated by the Adjusted Rand Index \cite{rand1971objective} and is done by comparing the cluster label $c_i$ obtained with the ground truth $y_i$ labels related to the tumor subtype provided as clinical data. It is important to remark that the ground truth labels are never used to select the features and are only used to measure how well the k-means groups the tumor samples. It is computed as
\begin{equation}
    \text{Rand Index} = \frac{\text{A}+\text{B}}{\text{A}+\text{B}+\text{C}+\text{D}}
\end{equation}
where A is the number of tumor sample-pairs assigned to the same cluster and belonging simultaneously to the same tumor subtype, B is the number of tumor sample pairs assigned to different clusters and simultaneously  belonging to different tumor subtypes, C is the number of tumor sample pairs assigned to the same cluster but belonging to different tumor subtypes and D is the number of tumor sample-pairs assigned to different clusters and belonging to the same tumor subtype. The Rand Index can be thought as a clustering accuracy and takes values from 0 to 1 where a value close to 0 means a random and non informative clustering results regarding the ground truth clinical labels. When the Rand Index is close to 1 it means that almost every cluster is populated with tumor samples of the same subtype which is a desired score.

\subsection{Baseline methods}

To evaluate the performance of LKFS first we stablish a baseline from existing methods. In this work LKFS is compared with the Sparse K-Means (SKM) \cite{witten2010framework} and the Spectral Feature Selection (SPEC) \cite{zhao2007spectral}. Both have been designed to perform unsupervised feature selection for clustering.\\
The SKM method computes via an optimization problem feature weights $ \mathbf{w} = [w_1, ..., w_d]$ and applies a lasso-type penalty $||w||_1 < \alpha$ to select the most important features while doing k-means clustering. The feature weights are a measurement of the variable importance in clustering. The SKM method is based on the K-means type family of algorithms and assigns a larger weight to the features that have a smaller sum of intra-cluster distances and a smaller or zero weight to the features which a high intra-cluster distance. \\
The other benchmark method is the Spectral Feature Selection for unsupervised learning (SPEC) which is based on spectral graph theory. SPEC uses a pairwise similarity matrix $\mathbb{S}$ between samples to build a graph $\mathbb{G}$ where each node is a sample and each edge is the similarity measurement. The idea with SPEC is to select the features that are consistent with the graph structure. The objective of SPEC is to select features that gives similar values to samples that are near each other in the graph. A graph $\mathbb{G}$ can be built from the pairwise similarity obtained from $\textbf{X}$. Then the SPEC method makes a feature ranking based on the Normalized Cut of the graph $\mathbb{G}$ by using the corresponding Laplacian matrix from the graph. \\
All the presented methods have the number of features to select as an hyperparameter $p$.

\section{\uppercase{Results}}
Experiments on each dataset have been conducted ten times by randomly selecting $80\%$ of the data. At each random iteration and in the following order data pre-processing, feature selection and clustering are performed. Then the evaluation by RED and Rand Index is averaged among all random iterations. A set of different number of selected $p$ features is used where $p = [10,20,30,40,50]$. A set of $k$ clusters are obtained where $k = [2,3,4,5]$.\\  
The first evaluation to be done is the RED score for each subset of selected features $p$ of each method on each dataset. 

\begin{figure}[h]
  \centering
   \includegraphics[width=4.6in]{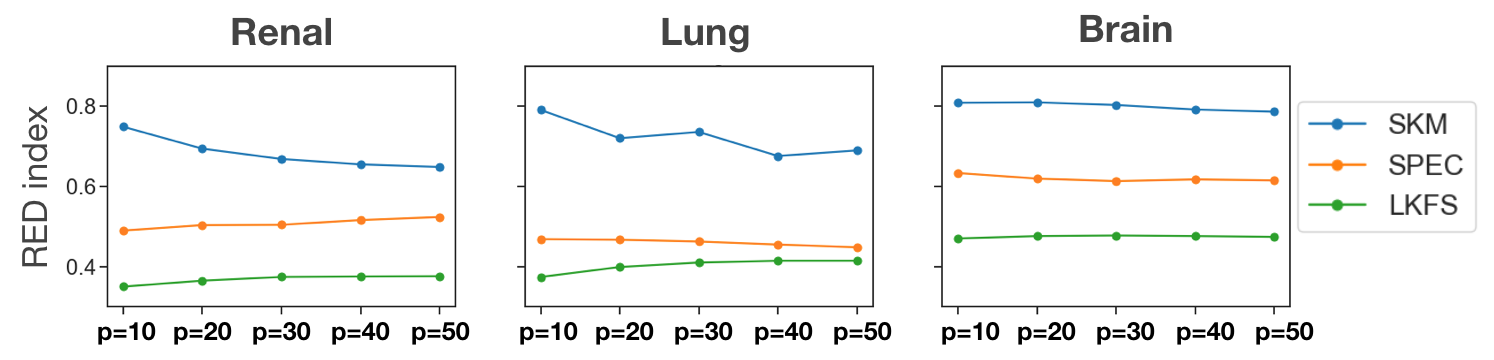} 
  \caption{RED index of the selected features on each dataset by each method.}
 \end{figure}

Figure 2 shows the results of the RED score. It is observed that LKFS has the lowest RED score in all the experiments followed by the SPEC method while SKM has the highest RED. This evidences that the features selected by the LKFS have the lowest redundancy which is a desired result since these are more informative and less redundant.

\begin{figure}[h]
  \centering
   \includegraphics[width=4.5in]{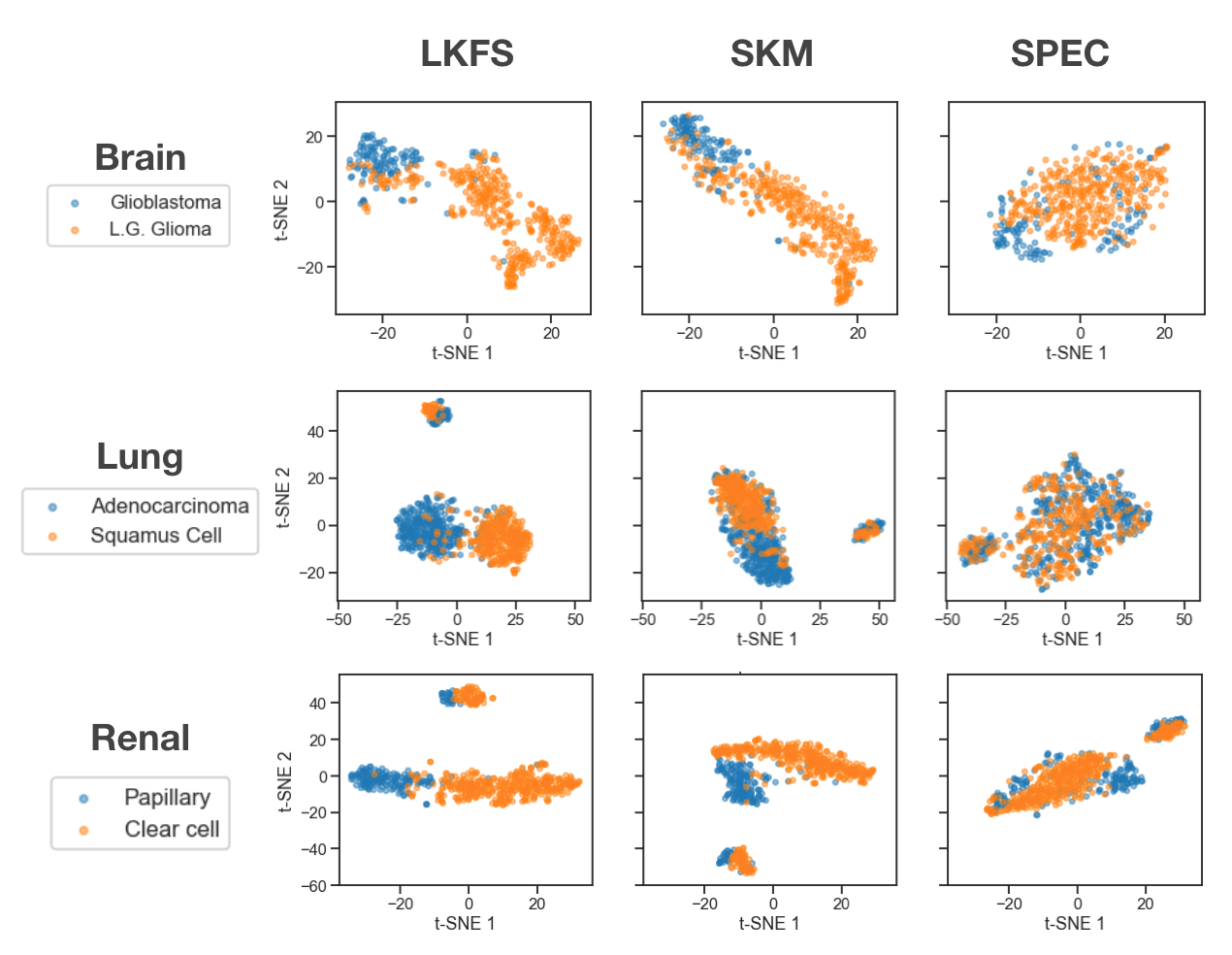} 
  \caption{Two dimensional t-SNE scatter plot for by dataset and method. Each tumor subtype on each dataset is highlighted in blue and orange respectively.}
 \end{figure}
 
Figure 3 shows a two dimension t-SNE scatter plot of each dataset by each method whith $p = 50$. For the Brain, Lung and Renal datasets LKFS shows how clusters tend to group tumors of the same subtype together in two main clusters. The SKM tend to polarize different tumor subtypes within the same cluster structure and finally SPEC fails to separate effectively the two tumor subtypes of each dataset.

\begin{figure}[h]
  \centering
   \includegraphics[width=4.5in]{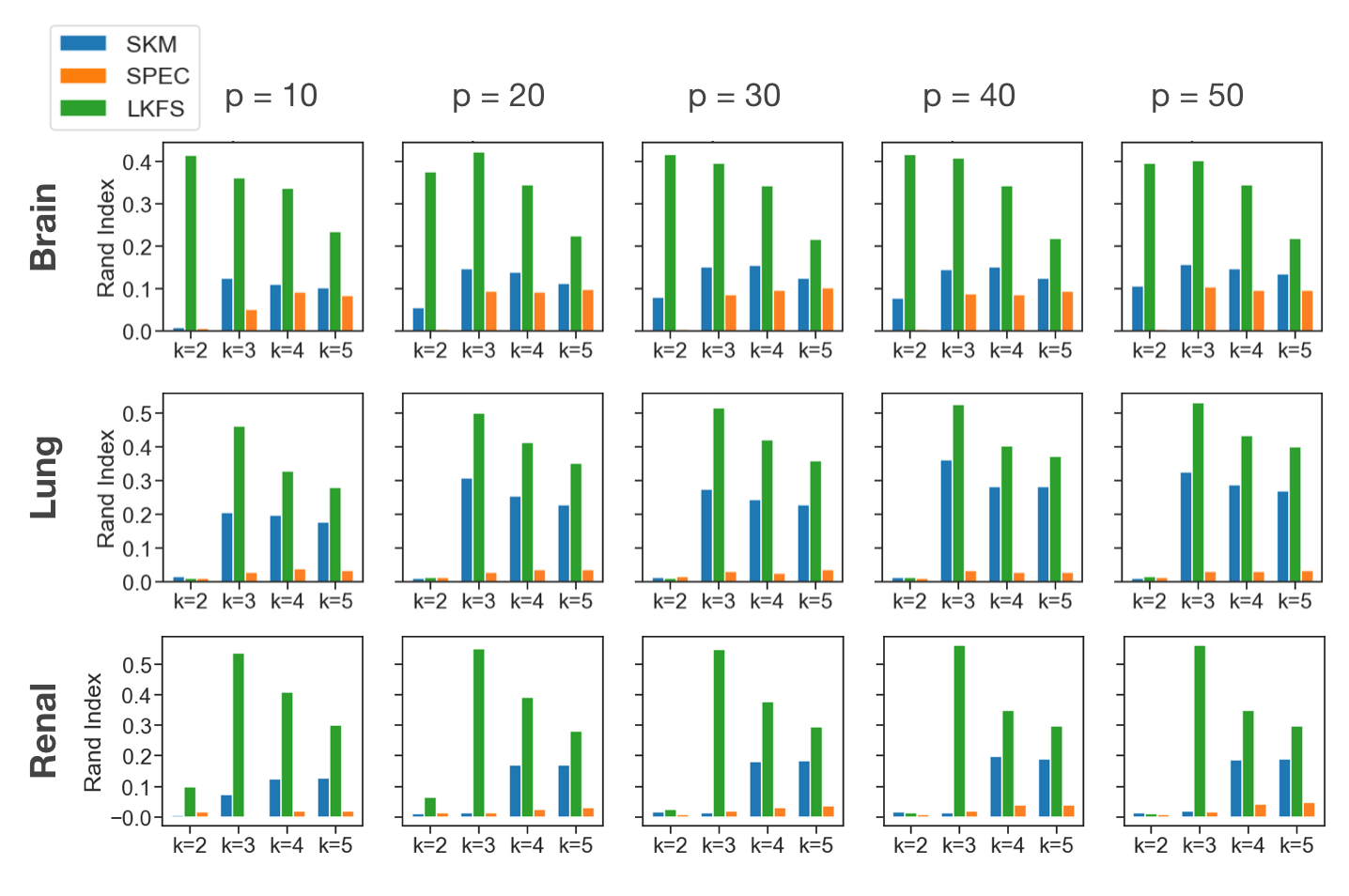} 
  \caption{Rand Index on each method and dataset for different values of $k$ clusters and $p$ features.}
 \end{figure}

Then K-means is applied on the selected features and the quality of clusters measured by the Rand-Index. Figure 4 shows the Rand-Index for different number of $k$ clusters on different number of $p$ selected features for each dataset and for each method. In the Brain dataset the LKFS method clearly outperforms the SPEC and SKM methods for every value of $p$ and $k$. In both the Renal and Lung datasets the LKFS outperforms the benchmark methods from $k = 3$ to $k = 5$. For these two datasets the three methods have a considerably low rand-index with $k = 2$ since in every case a small and isolated cluster composed by the two subtypes affects the performance and can be observed in Figure 3.

\subsection{Discussion}

This work proposes the LKFS method which selects by an unsupervised approach a reduced gene subset from more than 8.000 genes to less than 50 in three different types of cancer datasets. The selection of features is done by improving the alignment between a kernel $K_z$ obtained from the latent space $\mathcal{Z}$ learned via an autoencoder and by the resulting one $K_{\mu}$ after multiple kernel learning on feature-wise kernels. The approach is based on building a target representation of the input data with the latent space of an autoencoder. This target representation has a reduced level of noise and is low dimensional. LKFS selects only the features that align the most to the target representation by kernel alignment. The target representation can be interpreted as a prior distribution that guides the selection process.\\
To measure the quality of the unsupervised feature selection process the Redundancy score RED is computed on the selected features. In addition a K-means clustering method is applied and the cluster performance is evaluated by the homogeneity of ground truth labels across clusters by the Rand Index.\\ 
From the experimental results it is observed that LKFS outperforms the SKM and SPEC methods by Rand Index and RED score for almost all the datasets and number of selected features. This suggest that LKFS is able to selects low redundancy features from high dimensional input space that contributes to find well defined clusters composed mainly from one tumor subtype. In contrast, SKM and SPEC select features with higher redundancy which do not contribute enough to build separated clusters neither to group samples from the same tumor subtype together. \\ 
One of the advantages of LKFS comes from the target representation of the Autoencoder. This representation captures the salient features of the input dataset and then by MKL the selected features will capture approximately the same data structure. Finally, LKFS provides three outputs. The first is the subset of selected features which is considerably reduced in comparison with the original feature set. The second one is the latent space provided by the autoencoder. The latent space serves not only as a target representation for the MKL process but also as a tool for data exploration since it can summarizes in a lower dimensional space the salient features of the original data. The third output is a set of tumor clusters for further subtype discovery. \\
One limitation of LKFS relies in the need to train two models, the AE and the MKL. In this architecture the AE conditions the quality of the selected features. 

\section{\uppercase{Conclusions}}
\label{sec:conclusion}
This work proposes an unsupervised feature selection method named LKFS which can select a considerably reduced subset of meaningful and low redundant features from high dimensional gene expression data. Experimental results show that the proposed method outperforms two unsupervised feature selection algorithms by analyzing the quality of the clusters built on the selected features. LKFS has been evaluated on tumor gene expression datasets from Lung, Renal and Brain Cancer patients and select features that help to identify tumor subtypes without any supervised approach. For this reason LKFS is a useful model for pattern recognition and data mining in a variety of cancer types and high dimensional biological applications. Further work will include multi-omics data fusion approaches in order to consider genomic, proteomic or metabolomic features simultaneously.

\section*{Acknowledgment}
This work was supported by the doctoral program of the Universidad Tecnologica Nacional in Argentina, the UTN FRBA doctoral school of Signal Processing and the Universite de Technologie de Troyes. Also it was funded by grants from CONICET, ANPCyT and FOCEM-Mercosur.

%\section*{References}

\bibliographystyle{apalike}
{\small
\bibliography{ref}}

\end{document}